\documentclass{article}

\usepackage{arxiv}

\usepackage[utf8]{inputenc} 
\usepackage[T1]{fontenc}    
\usepackage{url}            
\usepackage{booktabs}       
\usepackage{amsfonts}       
\usepackage{nicefrac}       
\usepackage{microtype}      
\usepackage{lipsum}
\usepackage{graphicx} 
\usepackage{caption} 
\usepackage{appendix}
\usepackage{subfig}
\usepackage{chngpage}

\usepackage[ruled,vlined]{algorithm2e}
\usepackage{tabularx}

\usepackage{amssymb,amsmath, amsthm}
\usepackage[switch]{lineno}

\title{On Duality Gap as a Measure for Monitoring GAN Training}

\author{
  Sahil Sidheekh \\
  
  \texttt{2017csb1104@iitrpr.ac.in} \\

 \and
  \textbf{Aroof Aimen}\thanks{equal contribution} \\
  
  \texttt{2018csz0001@iitrpr.ac.in} \\

\and
   \textbf{Vineet Madan}\footnotemark[1] \\
  
  \texttt{2017csb1119@iitrpr.ac.in} \\

\and
   \textbf{Narayanan C. Krishnan}\thanks{corresponding author} \\
  
  \texttt{ckn@iitrpr.ac.in} \\
  
 \and
    Indian Institute of Technology, Ropar \\
    \\
 }

\newtheorem{definition}{Definition}
\begin{document}
\maketitle

\begin{abstract}
 Generative adversarial network (GAN) is among the most popular deep learning models for learning complex data distributions. However, training a GAN is known to be a challenging task. This is often attributed to the lack of correlation between the training progress and the trajectory of the generator and discriminator losses and the need for the GAN's subjective evaluation. A recently proposed measure inspired by game theory - the duality gap, aims to bridge this gap. However, as we demonstrate,  the duality gap's capability remains constrained due to limitations posed by its estimation process. This paper presents a theoretical understanding of this limitation and proposes a more dependable estimation process for the duality gap. At the crux of our approach is the idea that local perturbations can help agents in a zero-sum game escape non-Nash saddle points efficiently.  Through exhaustive experimentation across GAN models and datasets, we establish the efficacy of our approach in capturing the GAN training progress with minimal increase to the computational complexity. Further, we show that our estimate, with its ability to identify model convergence/divergence, is a potential performance measure that can be used to tune the hyperparameters of a GAN. \end{abstract}



\section{Introduction}
Generative Adversarial Network(GAN) \cite{goodfellow2014generative} is probably one of the most important inventions in recent years that is widely popular for modeling data distributions, especially for high dimensional data such as images. A GAN aims to learn a data distribution under adversarial gameplay between two agents - a Discriminator (D) and a Generator (G) parameterized by $\theta_d$ and $\theta_g$ respectively. The discriminator ($D: X \rightarrow [0, 1] $) differentiates between real data samples coming from the distribution to be learned and fake data samples generated by the generator, while the generator ($G: Z \rightarrow X $) aims to fool the discriminator; where $X$ is the data space and $Z$ is the low dimensional latent space. Formally, the GAN objective is defined as 
\begin{equation}\label{eq:1}
    \underset{\theta_g\in \Theta_G}{\min} \ \underset{\theta_d \in \Theta_D}{\max} \ F(D,G) ,
\end{equation} 

\begin{equation} \label{eq:2}
      F(D,G) = {\mathop{\mathbb{E}_{ \textbf{x} \sim P_{r}}}[\log(D( \textbf{x}))]   + \mathop{\mathbb{E}_{\textbf{z} \sim P_{z}}}[\log (1-D(G(\textbf{z})))]}
\end{equation}
where $P_{r}$ is the real data distribution and $P_{z}$ is a latent prior. The adversarial game using the above objective has been proven to minimize the Jensen-Shannon (JS) divergence between the true and generated data distributions.


Unlike classical machine learning tasks where the loss functions allow for direct inference on the training progress and a model's performance, GAN loss functions are non-intuitive and hard to interpret. This is partly due to the stochasticity involved in the training procedure, but mainly because the individual agents' loss functions also depend on the adversary's parameters. The alternate optimization changes the individual agents' loss surface at every iteration, making inference of training progress a challenging task.

Recently, Grnarova et al. suggest the use of duality gap as a measure to monitor the training progress of a GAN. They establish through exhaustive experimentation and theoretical reasoning that the duality gap, being a domain agnostic measure can be used to evaluate GANs over a wide variety of tasks. While the true duality gap is an upper bound on the JS divergence between the real and generated distributions, estimating the true duality gap is an intractable problem. They suggest a gradient descent approach to approximate the duality gap. It is shown that the approximated duality gap can provide insights into the performance a GAN.

While the idea of using the duality gap is promising, the approach to estimating it has an inherent limitation due to the nature of gradient descent and the GAN loss surface. Gradient descent seeks first-order stationary points \cite{jain2017}. In a GAN, a first-order stationary point can be a Nash or non-Nash critical point. We show that the duality gap computed through gradient descent cannot distinguish between these two critical points, which is fundamental to GAN training. Thus, it is not useful for monitoring GAN training. 

We propose an effective way to compute the duality gap using locally perturbed gradient descent. The central idea of the approach is to locally perturb a first-order stationary point before applying gradient descent. It can be shown that the local perturbation allows gradient descent to escape from a non-Nash critical point. This effectively differentiates the duality gap computed at Nash and non-Nash critical points.

Overall, we make the following contributions
\begin{itemize}
    \item We demonstrate a problem with the prevalent approach to estimate the duality gap.
    \item We propose a theoretically grounded approach to estimate the duality gap through local perturbations that overcomes the limitations of the earlier approach.
    \item We conduct extensive experiments on a wide variety of GAN models and datasets to demonstrate the domain and model agnostic nature of the proposed measure to monitor the training process.
    \item We demonstrate that the duality gap can be used as a measure to influence the training process of a GAN, such as tuning hyperparameters. We also show that controllers that maintain the delicate balance between the generator and discriminator updates can be learned using rewards based on duality gap.
\end{itemize}


\section{Related Work}
GANs have come a long way since it was proposed by \cite{goodfellow2014generative}. The various challenges associated with a GAN such as convergence to non-Nash points, lack of diversity in generated samples, diminishing gradients, and enforcing optimal balance between the generator and discriminator have been well-studied. Several loss functions and model architectures have been proposed in this regard \cite{li2017mmd, arjovsky-chintala-bottou-2017, gulrajani2017improved, DBLP:conf/iclr/MiyatoKKY18,DBLP:journals/corr/abs-1901-00838, heusel2017gans,DBLP:journals/corr/RadfordMC15,zhang2019progressive}. However, despite the numerous advantages, they still fail to provide an insight into the training progress of a GAN, owing to the non-intuitive nature of the associated loss curves. Thus, developing efficient measures for evaluating and monitoring GANs has attracted much interest in recent years \cite{DBLP:journals/corr/abs-1802-03446, lucic2018gans,DBLP:journals/corr/abs-1808-04888}. 

While log-likelihood has been used traditionally to train and evaluate generative models, it is not a suitable measure for a likelihood-free model like GAN. Further, estimating likelihood is intractable and subject to error in higher dimensions \cite{theis2015note}. Among the popular metrics to assess GANs are - Inception Score (IS)\cite{salimans2016improved} and Frechet Inception Distance (FID) \cite{heusel2017gans}. IS evaluates a batch of generated images based on the ability of a pre-trained inception network to accurately classify them among known Image-Net classes - while the entropy of the predicted labels of the generated distribution captures the diversity, quality is implicitly captured by the inception score as the proximity of the generated samples to the real samples. However, as IS does not statistically compare the real and generated distributions, it is not sensitive to mode collapse. FID overcomes this weakness of IS, but it assumes the embedding from the inception network to be Gaussian, which is unrealistic. Both FID and IS are restricted only to images. \cite{sajjadi2018assessing} propose precision and recall to capture the fidelity and diversity aspects of generative models. Precision quantifies the quality of the generated samples, and recall estimates the proportion of the real distribution covered by the generated distribution. However, they assume that the embedding space is uniformly dense and rely on the sensitive $k$-means clustering to determine the support set. \cite{kynkaanniemi2019improved} improve precision and recall by estimating the probability density using $k$-nearest neighbors. However, the improved precision and recall are sensitive to outliers. \cite{tolstikhin2017adagan,DBLP:journals/corr/abs-2002-09797} propose Density and Coverage to reduce the overestimation of manifold due to outliers. 

Apart from the requirement of labeled data (pre-trained networks), the above mentioned measures are domain-dependent, constraining their application to image datasets. Further, the best values of these measures are subjective to the real data, and hence there is no generic bound on these metrics to judge that the training has converged. There is a need for an intuitive metric that is domain agnostic, computationally feasible, and gives insight into the GAN training. \cite{grnarova2019domain} propose such an efficient metric - the Duality Gap to monitor the training progress of GANs. Unlike other metrics, duality gap is neither restricted to a specific data distribution nor requires a pre-trained neural network and is lower bounded by JS divergence between real and generated distributions. Our work is a study of the estimation process of the duality gap, aiming to improve its efficiency as a performance monitoring tool for GANs.

\section{Methodology}
\subsection{Preliminaries}
\textbf{Game Theoretic and Optimization Perspectives : }  
In the standard formulation of a GAN (equation 1), the equilibrium for the adversarial game play between the generator and the discriminator is the Nash equilibrium \cite{goodfellow2014generative}. At such an equilibrium, the divergence between the real and generated distribution is minimum. Formally, a configuration or strategy $(\theta_g^*, \theta_d^*)$ is said to be a pure Nash equilibrium, if:
\begin{equation} \label{eq:3}
    F(\theta_g^*,\theta_d) \leqslant F(\theta_g^*,\theta_d^*) \leqslant F(\theta_g ,\theta_d^*) \ \forall \ \theta_g,\theta_d
\end{equation}
Equivalently,
\begin{equation} \label{eq:4}
      \underset{ \theta_g\in \Theta_G}{\min} F(\theta_g,\theta_d^*) \ = \ \underset{\theta_d \in \Theta_D}{\max} \ F(\theta_g^*,\theta_d) = F(\theta_g^*,\theta_d^*)
\end{equation}

However, finding the global Nash equilibrium is hard because the loss surface for the GAN optimization is not convex-concave \cite{jin2019local}. Typically, a Local Nash Equilibrium (LNE) is what we expect to attain. Formally, a configuration $(\theta_g^*, \theta_d^*)$ is said to be a pure LNE, if 
$ \exists \delta > 0 \ \mbox{such that} \ \forall \ \theta_g,\theta_d \ \mbox{ with } \ \| \theta_g - \theta_g^* \| \ < \delta \ \mbox{and} \ \| \theta_d - \theta_d^* \| < \delta $, 
\begin{equation}
 F(\theta_g^*,\theta_d) \leqslant F(\theta_g^*,\theta_d^*) \leqslant F(\theta_g ,\theta_d^*) 
\end{equation}
It is known that the GAN loss surface has abundant saddle points \cite{DBLP:journals/corr/abs-1901-00838,choromanska2015loss}. Thus we must understand the characteristics of the LNE, a saddle point that we seek. As neural networks parameterize our models, we use gradient-based alternate optimization to seek the LNE $(\theta_g^*, \theta_d^*)$, which is a local maximum w.r.t the maximizing agent (discriminator) and a local minimum w.r.t the minimizing agent (generator). Formally, the local Nash Equilibrium $ (\theta_g^*, \theta_d^*)$ satisfies
\begin{equation}
    \nabla_{\theta_g} F(\theta_g^*, \theta_d^*) = 0  \text{ and } \nabla_{\theta_d} F(\theta_g^*, \theta_d^*) = 0 
    \label{eq:non-nash}
\end{equation}
\begin{equation}
    \frac{\partial^2 F(\theta_g^*, \theta_d^*)}{\partial {\theta_g}^2} \succ 0 \text{ and } \frac{\partial^2 F(\theta_g^*, \theta_d^*)}{\partial {\theta_d}^2} \prec 0, 
\end{equation}
We expect the gradient dynamics associated with the alternate optimization to enforce convergence to a local Nash Equilibrium. However, in practice, there are a plethora of non-Nash critical points that are attractors (satisfy only equation \ref{eq:non-nash})  under the gradient dynamics \cite{DBLP:journals/corr/abs-1901-00838}, thus making the training of GANs unstable and cumbersome.

 \subsection{Duality Gap for Monitoring GAN Training}
 A good measure for monitoring GAN training should be domain agnostic and enable easy inference of the training progress. Duality Gap \cite{grnarova2019domain} is a recently proposed measure motivated by principles of game theory.
 
 \begin{definition}(Duality Gap) Let $\Theta_D $ and $\Theta_G$ denote the parameter spaces of the discriminator and the generator respectively. Then for a pure strategy $(\theta_g,\theta_d)$ such that $\theta_g \in \Theta_G$ and $\theta_d \in \Theta_D$, the \textit{Duality Gap}, is defined as 
 
 \begin{equation}\label{eq:5}
 DG(\theta_g,\theta_d) =  \underset{\theta_d^{'} \in \Theta_D}{\max} \ F(\theta_g,\theta_d^{'} )  \ - \ \underset{ \theta_g^{'} \in \Theta_G}{\min} F(\theta_g^{'},\theta_d)
 \end{equation}
 \end{definition}
 
 Intuitively, the duality gap measures the maximum payoff the agents can obtain by deviating from the current strategy. At Nash Equilibrium, as no agent can unilaterally increase their payoff, the duality gap is zero. This is implied by equation \eqref{eq:4}. Further, the duality gap is always non-negative. Grnarova et al. show that the duality gap is lower bounded by the Jensen-Shannon divergence between the true and generated data distributions. While the duality gap seems to be a useful performance measure for GANs, computing the exact duality gap at any point is an NP-hard problem as it involves finding the extrema of non-convex functions \cite{thekumparampil2019efficient,grnarova2019domain}.
 
 \subsubsection{Duality Gap Estimation}
The duality gap is estimated through an iterative gradient based optimization \cite{grnarova2019domain}. An auxiliary generator and discriminator are initialized to the current values $(\theta_g^t,\theta_d^t)$ of their GAN counterparts at iteration $t$. The auxiliary discriminator is optimised for a fixed number of iterations to obtain the worst case discriminator ($\theta_d^w$) to compute $M_1 = F(\theta_g^t,\theta_d^w)$. Analogously, the auxiliary generator is optimised to obtain the worst case generator ($\theta_g^w$) to compute $M_2 = F(\theta_g^w,\theta_d^t)$. The difference $M_1 - M_2$ is the estimated duality gap at the iteration $t$.

\subsubsection{The Challenge }
This estimation process seems very intuitive from the game theory perspective as it imitates the individual players' efforts to unilaterally increase their payoffs by deviating from their \textit{current strategy}. However, it limits the ability of the estimated duality gap as a performance metric to distinguish clearly between stable mode collapse, divergence and convergence encountered during GAN training. This is because, the gradient based optimizations for computing $M_1$ and $M_2$ seek first order stationary points i.e. points where the gradients of the objective function  w.r.t to the optimizing agent is zero. As Nash and non-Nash critical points are both first order stationary points \cite{fiez2019convergence}, the duality gap estimated would be very close to zero. This is evident from the updates for estimating the worst case generator and discriminator. At the critical point $(\Tilde{\theta}_g, \Tilde{\theta}_d)$, $\nabla_{\theta_g} F(\Tilde{\theta}_g, \Tilde{\theta}_d) \mbox{=} 0$ and  $\nabla_{\theta_d} F(\Tilde{\theta}_g, \Tilde{\theta}_d) \mbox{=} 0$ due to the first-order stationarity property. Thus it acts as an attractor in the gradient optimization for estimating the worst case generator ($\theta_g^w$) and discriminator ($\theta_d^w$). Thus $({\theta_g^w}, {\theta_d^w}) \mbox{=} (\Tilde{\theta}_g, \Tilde{\theta}_d)$, which in turn results in $M_1 \mbox{=} M_2$ and hence $DG(\Tilde{\theta}_g, \Tilde{\theta}_d) \mbox{=} 0$. Thereby making it hard to differentiate between Nash and non-Nash critical points. However, at a non-Nash critical point, at least one of the agents can increase the payoff by deviating from the current strategy. It is the inherent limitation of the estimation process that the auxiliary models are unable to escape the non-Nash critical points \cite{DBLP:journals/corr/abs-1901-00838}.


To circumvent the optimization for estimating the worst discriminator and generator,  \cite{grnarova2019domain} suggests estimating the approximate duality gap by choosing the most adversarial discriminator and generator from the saved set of snapshots of generator and discriminator parameters at different training time instants. This approximation requires the book-keeping of the discriminator and generator snapshots at different timestamps throughout the training. It is non-intuitive as it is uncertain if the worst generator/discriminator for a particular configuration would have been encountered previously during the GAN training.


 \subsection{Perturbed Duality Gap }
We propose an effective way to compute the duality gap accurately using locally perturbed gradient descent. The method introduces perturbations to the worst-case agent's initial point in its local neighborhood before applying gradient descent. Algorithm \ref{alg:Perturbed Duality Gap} summarizes the modified estimation process. The auxiliary generator and discriminators' parameters are initialized to the current iterate values, $\theta_g^t$ and $\theta_d^t$, respectively. The initialization is perturbed by adding noise sampled uniformly from a ball of radius $\sigma$. Optimization is performed on the perturbed initialization to estimate the worst-case generator ($\theta_g^{w}$) and discriminator ($\theta_d^{w}$). The success of the method relies on the perturbation that is applied to the initialization. We discuss the intuition of the local perturbation in the next subsection and present a way to limit the radius of the perturbation ball in the experiments.


\begin{algorithm}
\SetAlgoLined
\textbf{Input:} Current iterate values of G and D - $\theta_g^t$, $\theta_d^t$\\
$\theta_d^a \longleftarrow \ \theta_d^t + \delta$\\
$\theta_g^a \longleftarrow \ \theta_g^t + \delta$\\
where $\delta \sim U([-\sigma, \sigma])$ \\
\For{i in Iterations}{$\theta_d^a \longleftarrow \ \theta_d^a + \eta \nabla_{\theta_d^a} F(\theta_g^t , \theta_d^a)$ \\
$\theta_g^a \longleftarrow \ \theta_g^a - \eta \nabla_{\theta_g^a} F(\theta_g^a , \theta_d^t)$ \\}
$(\theta_g^{w}, \theta_d^{w})$ are the converged worst cases\\
$M_1 = F(\theta_g^t , \theta_d^{w})$\\
$M_2 = F(\theta_g^{w} , \theta_d^t)$\\
return $DG(\theta_g^t , \theta_d^t)$ = $M_1 - M_2$\\
\caption{Perturbed Duality Gap\ DG ($\theta_g^t$, $\theta_d^t$)}
\label{alg:Perturbed Duality Gap}
\end{algorithm}

\subsubsection{Intuitive and Theoretical Understanding}
The intuitive and theoretical understanding of the proposed method is inspired from the work on escaping saddle points during gradient descent  \cite{DBLP:conf/icml/Jin0NKJ17}. Let $(\theta_g,\theta_d)$ be a configuration at which we would like to compute the duality gap. Without loss of generality, let us consider the optimization to compute $M_2$. Let $(\theta_g^a,\theta_d)$ be the perturbed initial point and $(\theta_g^w,\theta_d)$ be the estimated worst case configuration. We can visualize $(\theta_g^a,\theta_d)$ as sampled from a perturbation ball ($B$) centered at $(\theta_g,\theta_d)$ with radius $\sigma$. The three possible situations at $(\theta_g,\theta_d)$ are (1) Gradients are sufficiently large; (2) Gradients are close to zero, but hessian is not positive semidefinite ($(\theta_g,\theta_d)$ is a first order stationary point); (3) Gradients are close to zero and hessian is  positive semidefinite ($(\theta_g,\theta_d)$ is a second order stationary point). In the first case, as $F(.)$ is smooth, classical gradient descent will reduce the function value.

The second and third cases are of specific interest because they are usually encountered during GAN training as convergence to non-Nash and Nash critical points respectively. If $(\theta_g,\theta_d)$ is a non-Nash critical point (the second case), we would like gradient descent starting from $(\theta_g^a,\theta_d)$ (the perturbed initialization) to escape the saddle point thereby minimising the function value. This is guaranteed by the theoretical results in \cite{DBLP:conf/icml/Jin0NKJ17} (Lemma 10). The intuition behind the theory revolves around understanding the geometry of the perturbation ball ($B$) around the saddle point. Let us denote by $B_S$ - the stuck region inside $B$ from which gradient descent would be unable to escape the saddle point. Jin et al. show that this region's volume is minimal compared to the perturbation ball's overall volume. Thus after adding the perturbation to $\theta_g$, the point $\theta_g^a$ has a small chance of falling in $B_S$ and hence will escape from the saddle point efficiently. The same reasoning can be analogously applied to the discriminator parameters while computing $M_1$. 
The direct inference from the above is that the perturbed duality gap estimate would be non-zero at non-Nash critical points  and close to zero in the vicinity of Nash critical points, thus enabling better differentiation between convergence/non-convergence scenarios.

The local perturbations to the initial point might not be very intuitive from the game theory perspective as it does not precisely measure the effect of deviating from the \textit{current strategy}. However, the estimation process is consistent because the perturbations are local (reachable) and only for the agent that is to be optimized. Hence should not significantly affect the optima to be attained as the loss surface remains unchanged. Incorporating this subtle modification will enable us to differentiate between convergence to Nash and non-Nash critical points. 

We know that an LNE is a locally stable and attracting critical point for both the agents. Thus, despite the local perturbations, during convergence to a local Nash Equilibrium $(\theta_g^*,\theta_d^*)$, the worst case discriminator while computing $M_1$ and the worst case generator while computing $M_2$ would ideally correspond to $(\theta_d^*$ and $\theta_g^*)$ respectively. Thus the duality gap would still be close to zero.
When the agents converge to a non-Nash critical point $(\theta_g^{'},\theta_d^{'})$, the original estimation process for $M_1$ and $M_2$ would result in the worst case discriminator and generator restricted to $(\theta_d^{'}$ and $\theta_g^{'})$ due to the lack of gradients in the vicinity of the saddle point. However, the introduction of local perturbations displaces the agent from the saddle point and provides the additional momentum required to escape the non Nash critical point, preventing the estimated duality gap from saturating to zero.

 \section{Experiments and Results}

\begin{figure}[t]
\centering
 \includegraphics[width = 0.72\linewidth]{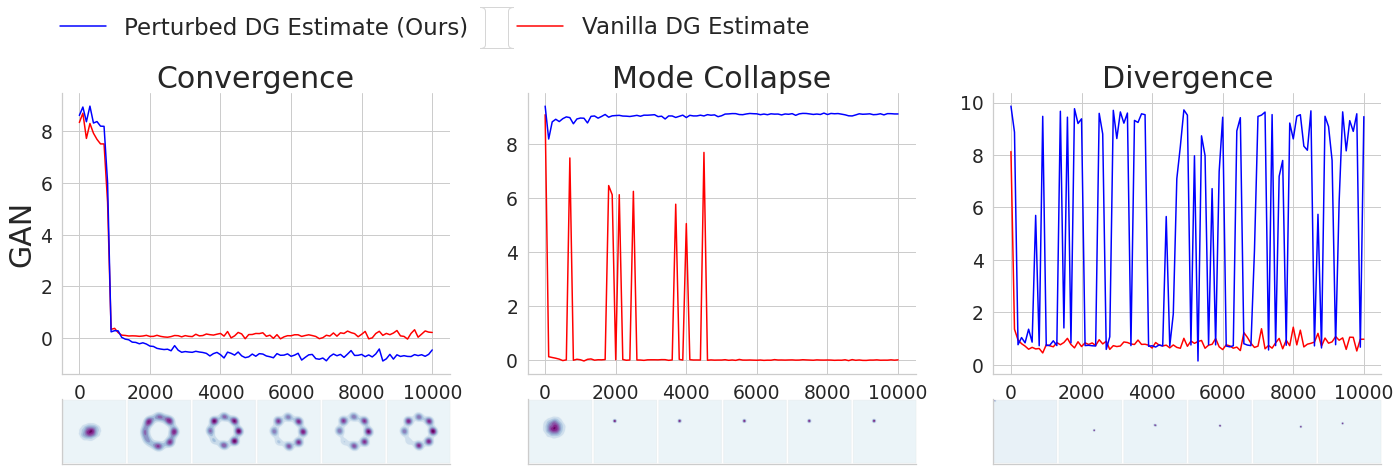}
  \includegraphics[width = 0.72\linewidth]{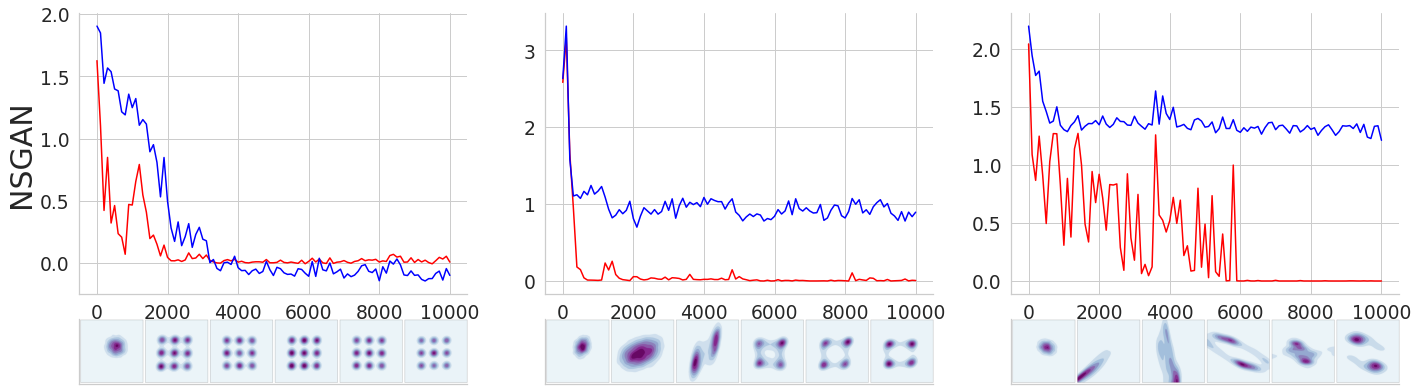}
  \caption{Perturbed and Vanilla DG estimate throughout the training progress of classic GAN (top row) and NSGAN (bottom row) on the RING and GRID datasets respectively for the scenarios - convergence (left column), mode collapse (middle column), and divergence (right column). For each sub-graph, X-axis corresponds to the number of iterations, and Y-axis corresponds to the estimated duality gap.}
\label{fig:2DGNS}
\end{figure}

We design experiments to investigate the commonly encountered failure cases during GAN training from the perspective of the duality gap (DG). We empirically show that the proposed method, which we refer to as perturbed DG estimate, is better equipped to monitor the GAN training than the estimate of Grnarova et al., which we refer to as vanilla DG estimate. Our objective is to only provide a rigorous comparison between the vanilla and perturbed DG estimate, for monitoring GAN training and is not on analyzing different GAN variants or datasets. The source code and other experimental details are publicly available \footnote{https://github.com/perturbed-dg/Perturbed-Duality-Gap}. We illustrate the hypothesized behaviour of the DG estimates near Nash and non-Nash critical points using a toy function in the supplementary material. We begin the discussion with the mixture of Gaussians. 
\subsection{Mixture of Gaussians}

\begin{figure}
\centering
\begin{tabular}{cc}
\subfloat[Classic GAN, DCGAN, NSGAN and WGAN-GP on MNIST dataset]{\includegraphics[width =0.65\linewidth]{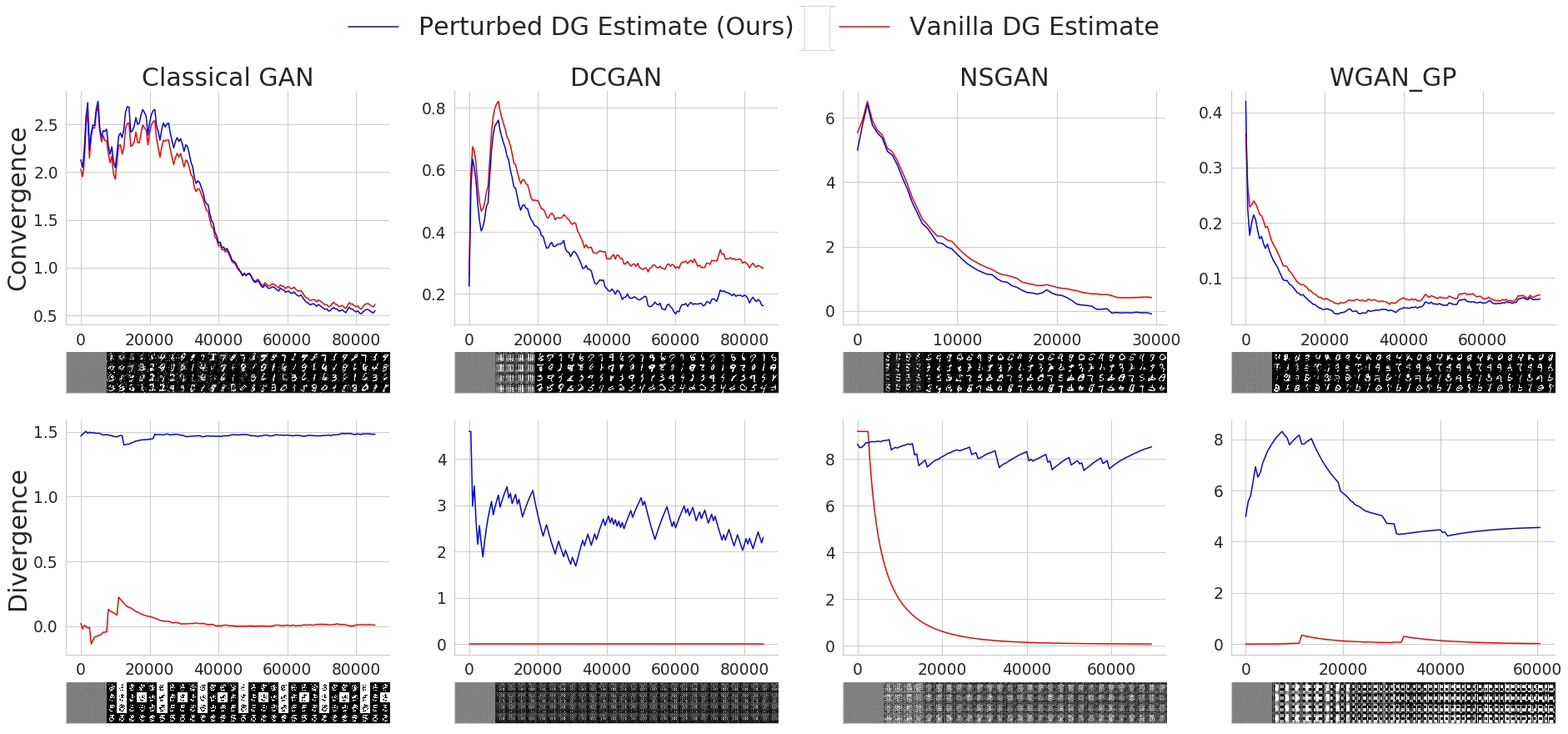}} &

\subfloat[WGAN-GP on CIFAR-10 and CELEBA  datasets]{\includegraphics[width =0.32\linewidth]{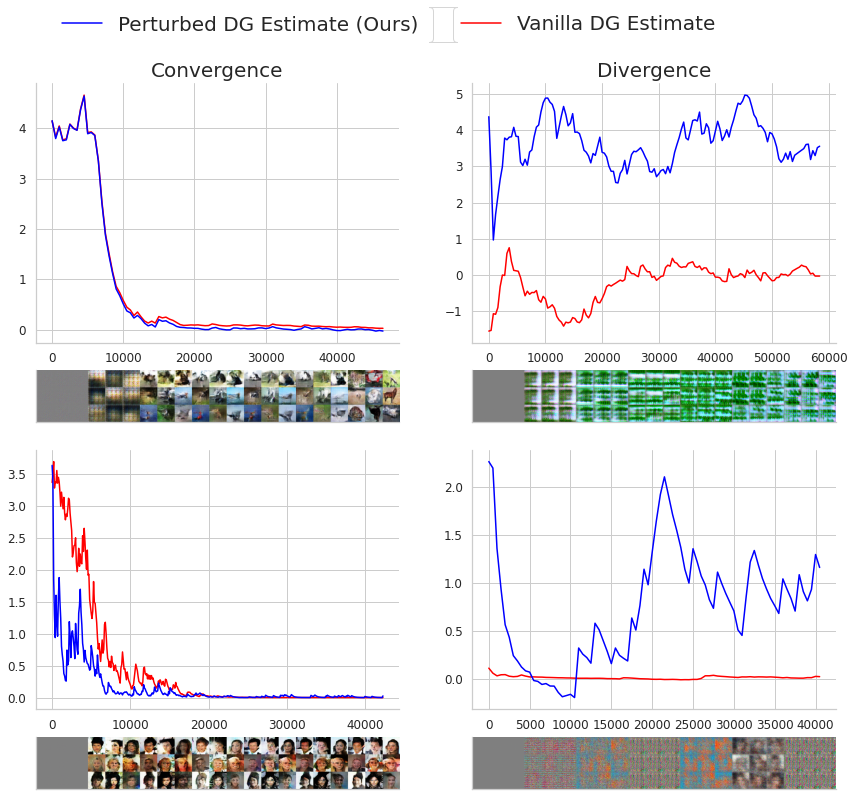}} 
\end{tabular}
\caption{Comparison of perturbed and vanilla DG estimate throughout the training progress of various GANs on image datasets across convergence and divergence settings.}
\label{fig:image_datasets}
\end{figure}

We use three toy Gaussian mixture datasets (RING, GRID, and SPIRAL) to monitor the GAN training process using the estimated duality gap. For each of these datasets, we train a classical GAN and its non-saturating (NSGAN) counterpart. We simulate convergence, mode collapse, and divergence scenarios by varying the learning rate and update sequence of generator and discriminator. We analyze the results for training the classical GAN and NSGAN on the RING  and GRID datasets. Similar patterns observed for the other datasets are discussed in the supplementary material.

Figure \ref{fig:2DGNS} demonstrates the sensitivity of the perturbed and vanilla duality gap estimates during convergence, mode collapse, and divergence throughout the training progress of a classic GAN and NSGAN on the RING and GRID datasets respectively. We observe that the behavior of perturbed and vanilla DG estimates is alike during convergence as both saturate close to zero as expected. The duality gap estimate being marginally negative at times despite the true duality gap being lower bounded by the JS divergence can be understood as a limitation posed by the approximation as discussed in \cite{grnarova2019domain}, and does not to a large extent affect its ability to capture the training progress. We observed that the Vanilla DG estimate saturates to values very close to zero during mode collapse and divergence while perturbed DG saturates (or oscillates) to non-zero positive values as expected. Thus, we verify that our approach can escape non-Nash saddle points and, therefore, can better track the training progress of the GANs.

\subsection{Image Datasets}


Having established the efficacy of our approach on synthetic datasets, we now focus on confirming the generality of perturbed DG across GAN architectures in learning higher dimensional data distributions. To this end, we train a classical GAN, DCGAN, DCGAN+NS (NSGAN), WGAN-GP over image datasets - MNIST, Fashion-MNIST, CIFAR-10, and CelebA \cite{deng2012the,DBLP:journals/corr/abs-1708-07747,cite-cifar10,conf/iccv/LiuLWT15}. To get an unbiased estimate of the DG, we split the datasets into a disjoint train, validation, and test sets to train the GAN, find the worst-case generator/discriminator, and evaluate the objective function w.r.t the worst-case agents respectively. We use classic GAN objective as a reference as suggested by \cite{grnarova2019domain} and train worst-case agents for 400 iterations using the optimizer of their GAN counterparts for estimating the duality gap. We simulate convergence and non-convergence scenarios by varying the learning rates.

 \subsubsection{MNIST and Fashion MNIST}
  
 Figure \ref{fig:image_datasets}(a) validates the generality of perturbed DG estimates across GAN architectures for the MNIST dataset. As expected, we observe that both vanilla and perturbed DG estimates saturate close to zero during convergence irrespective of the GAN architecture. However, the vanilla DG estimate saturates close to zero even when the GAN diverges. On the contrary, the perturbed DG estimate shows a higher sensitivity to the divergence setting by saturating to a non-zero positive value. We repeat this experiment on the Fashion MNIST dataset and illustrate the results in the supplementary material. In addition to convergence and divergence settings, the perturbed DG estimate is also sensitive to mode collapse even for complex data distributions as suggested by the results presented in the supplementary material.

 \subsubsection{CIFAR-10 and CelebA}

 We investigate the generality of perturbed DG for monitoring the GAN training progress on complex image distributions like CIFAR-10 and CelebA. We choose WGAN-GP, a widely accepted GAN model for learning high dimensional data distributions, for this experiment. Results of this experiment (Figure \ref{fig:image_datasets}(b)) show that the perturbed DG estimate can clearly differentiate between convergence and divergence, validating the hypothesis that the choice of the data distribution and GAN architectures do not constrain the sensitivity of the perturbed DG estimate towards convergence and (or) non-convergence. 

 \subsection{Influencing GAN Training using perturbed DG}

A generic framework for tuning the hyperparameters of a GAN with minimal human intervention has been an open problem in the community due to a lack of domain agnostic measures capable of accurately quantifying a GAN's performance. The convergence of a GAN largely depends on preserving the delicate balance between the generator and discriminator \cite{zhang2019progressive} that is governed by the tuning of hyperparameters. 
We show empirically that perturbed DG can be used to fine tune these hyperparameters. Further, perturbed DG can facilitate learning of meta-models that can automatically drive a GAN to convergence.
\subsubsection{Hyperparameter Search}

\begin{figure}
\centering
\begin{tabular}{cc}
\subfloat[]{\includegraphics[width = 0.32\linewidth]{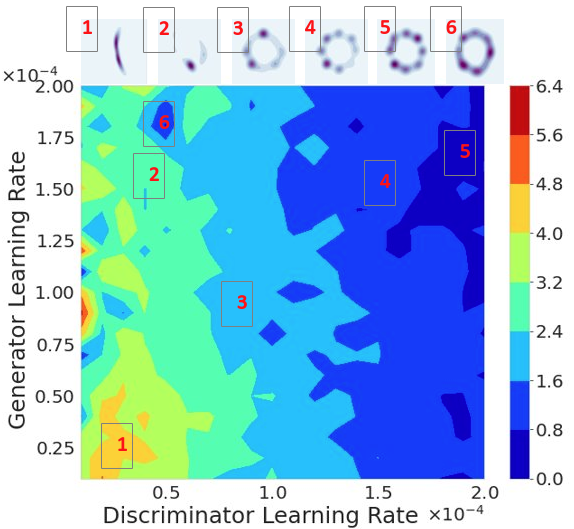}} &
\subfloat[]{\includegraphics[width = 0.32\linewidth]{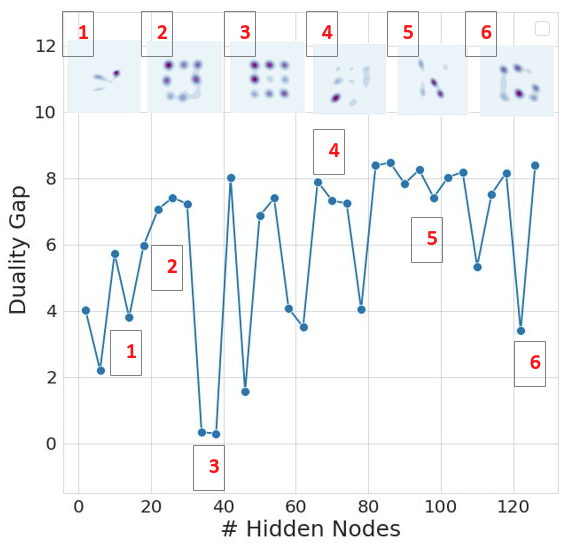}} 
\end{tabular}
\caption{[Best viewed in color] Perturbed DG trend across: (a) Learning rates (b) Model Capacity}
\label{fig:hyperparam_tune}

\end{figure}

We focus on tuning the learning rates and model capacity. We perform a grid search on these hyperparameters and compute the perturbed DG estimate at every choice to identify the optima. The learning rates of both the generator and the discriminator impact each other. Thus we search over the 2D space defined by both the parameters. 
Figure \ref{fig:hyperparam_tune}(a) depicts the contour plot for the perturbed DG estimate over this space for a classic GAN trained on the RING dataset. The optimal range of learning rates for the generator and the discriminator can be easily identified as the region where perturbed DG approaches zero. The optimal learning rates, as suggested by the perturbed DG estimate, are further verified by visualizing the similarity of the generated and true data distribution in these regions. As for model capacity, we search over the space defined by the number of hidden nodes in each layer of the models - we use the same number of layers and hidden nodes for both the generator and the discriminator.  This is visualized in Figure \ref{fig:hyperparam_tune}(b) for a classical GAN trained on the GRID dataset. We observe an optimum capacity for the models at which the game is balanced, and the perturbed DG estimate can identify it as the region where the duality gap is close to zero.
\subsubsection{Dynamic Scheduling}

\begin{figure}[t]
\centering
\begin{tabular}{cc}
\subfloat[GRID Dataset]{\includegraphics[width=0.40\linewidth]{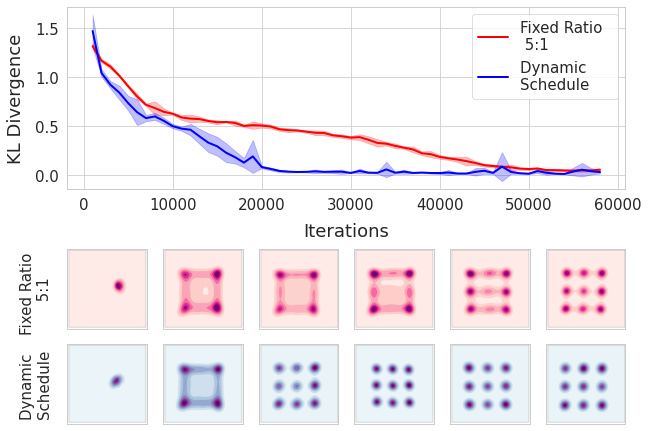}}
\subfloat[MNIST Dataset]{\includegraphics[width=0.40\linewidth]{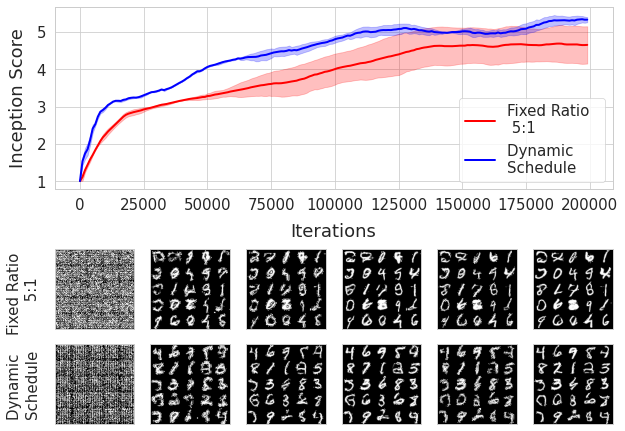}}
\end{tabular}
\caption{[Best viewed in color] Dynamic scheduling using perturbed DG based controller leads to faster convergence.}
\label{fig:autoloss_grid_mnist}
\end{figure}

Another dimension governing the balance between the generator and discriminator is the schedule that drives the underlying alternate iterative optimization. \cite{xu2018autoloss} shows that a dynamic schedule that considers the current state and the history of the optimization procedure over following a fixed schedule can lead to better convergence for GANs. They use a controller network, learned using a policy gradient method to maximize the expected reward defined in terms of the final performance of the GAN, to generate a dynamic schedule. However, they quantify the GAN performance using the Inception Score, limiting its applicability to only images. We extend their approach by superseding the inception score with perturbed DG as the reward for training the controller. We define the input space of the controller as a 4-tuple - (a) the log-ratio of the magnitude of the gradients of the generator and discriminator, an exponential moving average of - (b) generator's loss, (c) discriminator's loss, and (d) perturbed DG. Such a state representation is independent of the GAN architecture and captures sufficient information needed to infer the balance between the agents. As there is an inverse relation between perturbed DG ($DG$) and convergence of a GAN, we specify the reward as $\frac{\alpha}{DG + \epsilon}$, where $\alpha$ is a reward constant and $\epsilon$ is added for numerical stability. The domain agnostic nature of the duality gap enables learning dynamic schedules for GANs irrespective of the data distribution and also facilitates learning a domain agnostic controller that is generalizable across GAN architectures and domains. We verify this hypothesis by training a controller using perturbed DG to enforce convergence of a WGAN-GP on the 2D RING dataset. We then use this trained controller to drive the training of a WGAN-GP on the 2D GRID dataset and a convolutional WGAN-GP on the MNIST dataset.

The implementation details regarding the training procedure of the controller are provided in the supplementary material.
Figure \ref{fig:autoloss_grid_mnist} depicts the average performance of the dynamic schedule against the suggested discriminator-generator update ratio of 5:1 for WGAN-GP, in terms of domain-specific evaluation criteria for each the above settings - KL divergence for 2D GRID and Inception Score for MNIST. We observe that the dynamic schedule leads to faster and better convergence for each of the above settings. The KL divergence between the real and generated distributions quickly saturates close to zero for the 2D GRID and the higher inception scores for MNIST are obtained when using the dynamic schedule predicted by the controller network. Thus, while the balance enforced by the controller helps transcend manual tuning of the schedule, its generalizability, owing to the domain agnostic nature of perturbed DG, outweighs the effort invested in its training process.
 \subsection{Ablation Studies}

\begin{figure}[t]
    \centering
    \begin{tabular}{cc}
    \subfloat[]{\includegraphics[width = 0.40 \linewidth]{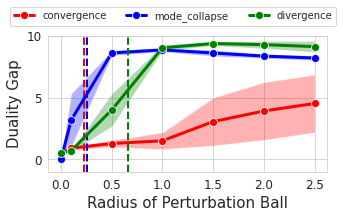}}
    \subfloat[]{\includegraphics[width = 0.40\linewidth]{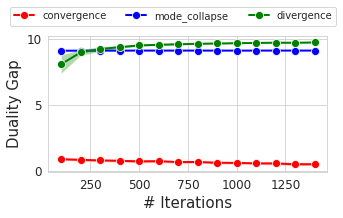}} 
    \end{tabular}
    \caption{Ablation Studies: (a) Perturbed DG estimate for varying radii of the perturbation ball. (b) Perturbed DG estimates w.r.t training iterations of the auxiliary GAN.}
    \label{fig:ablation-per}
\end{figure}

\subsubsection{Perturbation Ball Radius}
An unknown in the proposed perturbed DG estimation process is $\sigma$, the radius of the perturbation ball. We investigate the effect of $\sigma$ on DG estimate for three settings, namely convergence, mode collapse, and divergence. We train a classic GAN on the RING dataset for conducting this study. We estimate the saturated DG value at the end of 20,000 iterations for each setting and monitor the impact of $\sigma$ on the perturbed DG values across 50 trials. We observe from figure \ref{fig:ablation-per}(a) that the DG steadily increases with $\sigma$ during the convergence setting. However, there is a steeper increase in DG values for mode collapse and divergence setting saturating to a positive value. The size of the stuck region in the perturbation ball could explain this behavior. During convergence, the model converges to a Nash point enclosed in a larger stuck region within the perturbation ball due to which sufficiently large $\sigma$ is required to push the model off the stuck region. On the contrary, during non-convergence, the model converges to a non-Nash critical point enclosed in a comparatively thin stuck region. Therefore a small increase in $\sigma$ is sufficient to push the model off the stuck region.

We also notice an increasing trend in the DG variance for the convergence setting across trials compared to other scenarios. This behavior is explained by observing that the perturbations across the trials are stochastic, and for some trials, it may be sufficient for the model to escape the stuck region compared to other trials. However, even a small perturbation is sufficient for the model to escape the stuck region for mode collapse and divergence settings, further indicating that the stuck region's volume is small for these non-Nash critical points arrived during GAN training.

We now draw the attention towards the approximation of $\sigma$ by the standard deviation of the model's weights. In figure \ref{fig:ablation-per}(a), red, blue, and green vertical lines correspond to the average standard deviation of the model's weights for convergence, mode collapse, and divergence settings, respectively. It is evident from the figure that for perturbation ball radii comparable to these standard deviations, perturbed DG is able to differentiate the corresponding scenarios. 
\subsubsection{Computational Complexity}
The two contributing factors for an increase in the computational complexity while monitoring GAN training using perturbed DG are - number of training iterations of the auxiliary model and the frequency of perturbed DG computation. We investigate the additional cost imposed by the DG estimation process under these two factors.
    \begin{figure}[t]
    \centering
    \begin{tabular}{cc}
    \subfloat[]{\includegraphics[width = 0.30\linewidth]{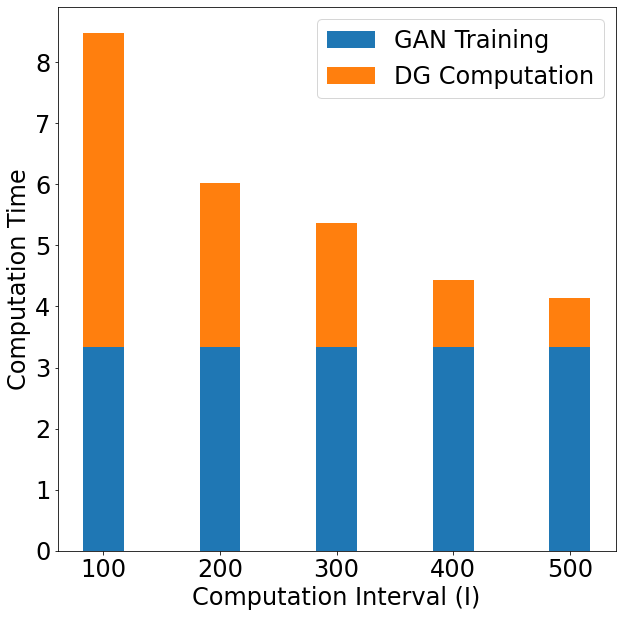}} &
    \subfloat[]{\includegraphics[width = 0.30\linewidth]{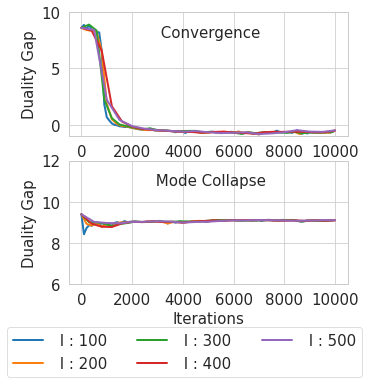}}
    \end{tabular}
    \caption{Computational Complexity (a) Overhead (average time taken per 100 training steps in seconds) of DG estimation at different computation intervals. (b) Perturbed DG captured at different intervals across GAN training process.}
    \label{fig:computational_complexity}
    \end{figure}
We train a classical GAN on the RING dataset. In the first experiment, we track the estimated DG values as a function of auxiliary GAN's training iterations. Figure \ref{fig:ablation-per}(b) shows the trend in the perturbed DG values. It is evident from the figure that less than 200 iterations are sufficient for the convergence of the DG for various settings. 
Figure \ref{fig:computational_complexity}(a) presents the increase in the computational time (averaged over 5 trials)  when estimating the DG at various intervals. The number of iterations for training the auxiliary GAN is fixed at 200 as suggested by the early stopping criteria. It can be seen that for higher computation intervals, which are quite practical to work with, the computation time is comparable to the GAN training process without any DG computation. We show that this higher DG computation interval does not degrade the trend captured by the duality gap in figure \ref{fig:computational_complexity}(b). The variance in the DG estimates across the iterations for varying estimation frequency is negligible.

\section{Conclusion}
Evaluating the training progress of a GAN has been an open problem in the machine learning community. Recently Grnarova et al. establish that the true duality gap is an upper bound on the JS divergence between the real and generated distributions, and through exhaustive experimentation, demonstrate that the duality gap is domain agnostic and can be used to evaluate GANs over a wide variety of tasks. Computing the true duality gap for a GAN is an intractable problem, and thus Grnarova et al. suggest a gradient descent approach to estimate its approximation. This paper shows that the duality gap estimation process suffers from an inherent limitation of being unable to distinguish between Nash and non-Nash attractors of the gradient dynamics associated with the GAN optimization. We present a thorough study of this limitation and propose a method to overcome it without increasing the computational complexity, through local perturbations. Through rigorous experiments, we demonstrate the usefulness of the perturbed duality gap for monitoring GAN training progress. We also show that it is possible to learn generic meta-models capable of driving GANs to convergence using the perturbed duality gap. This potentially opens up new possibilities for automated GAN training.

\bibliographystyle{unsrt}  
\bibliography{reference} 

\end{document}


\maketitle



\section{Overview}

This section details the experimental setup and observations left out in the main paper due to space constraints. The supplementary material is organized as follows: 

\begin{itemize}
  
  \item \textbf{GAN losses are non-intuitive:} We illustrate situations often encountered during GAN training, where the non-intuitive nature of the GAN loss curves are apparent. This motivates the need for a reliable metric for monitoring GAN training.
  

  \item \textbf{Toy function:} This section illustrates the behavior of perturbed and vanilla DG estimates in the vicinity of Nash and non-Nash critical points of a toy function.
  
  \item \textbf{Synthetic 2D datasets:} In this section, we compare the performance of vanilla and perturbed DG estimate during the training of various GANs (classical and NSGAN) on 2D synthetic datasets (RING, GRID, SPIRAL).   
  
  \item \textbf{Image datasets:} The comparative analysis of vanilla and perturbed DG on high dimensional image datasets is presented in this section.
 
  \item \textbf{Dynamic Scheduling Using Perturbed DG:}
     The implementation details and observations pertaining the training of a meta-model to drive a GAN convergence are presented in this section.
     
  \item \textbf{Ablation studies:} The last section analyzes the trend of the perturbed DG during GAN training w.r.t the radius of the perturbation ball $\sigma$.
\end{itemize}

\section{Non-Intuitive Nature of GAN Loss Curves}

\begin{figure}
        \centering
         \includegraphics[width = \linewidth]{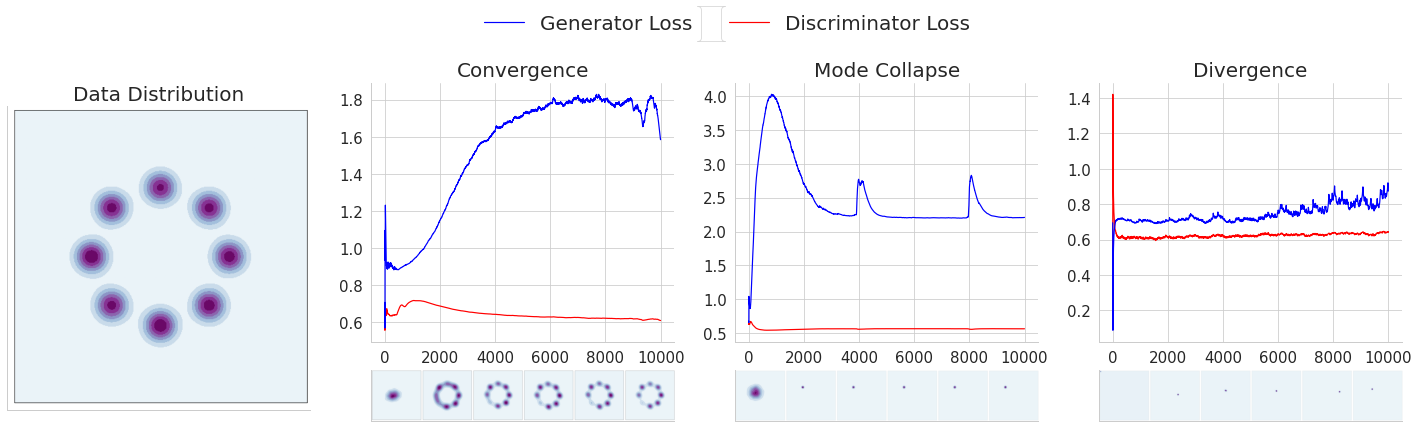}
          \caption{Generator and discriminator losses throughout the training progress of classic GAN on RING dataset for convergence, divergence and mode collapse settings.}
        \label{fig:gan_loss}
\end{figure}

With the advent of deep learning, most machine learning tasks have been reduced to function optimization problems, popularly solved through gradient descent iteratively. The surface of the optimization objective is hard to visualize due to the large number of parameters. However, for classical machine learning tasks, it is usually possible to infer whether the model has converged by analyzing the trend in the values of the objective function that is being optimized. We would ideally expect the loss curves associated with the model to decrease and eventually saturate, indicating convergence. Such intuitive inferences, however, cannot be made for a GAN. This is because GANs involve, in addition to the stochasticity, an alternating iterative optimization where the loss surfaces of the generator and discriminator change at every iteration. Thus, while we do expect the divergence between the real and generated distributions to eventually decrease, the losses of the individual models need not always decrease steadily to result in convergence \cite{DBLP:conf/iclr/FedusRLDMG18}. This is also implied by the adversarial nature of the GAN game - as the objectives of the individual agents are conflicting, an equilibrium cannot be attained through consistent reduction in the losses of both the models. Figure \ref{fig:gan_loss} shows the loss curves of a classic GAN trained over the RING dataset across different settings. We observe that during convergence, the discriminator's loss decreases, while the generator's loss increases where-after they saturate. However, similar behaviour is also observed in the loss curves during a stable mode collapse and divergence. In fact, during divergence, despite the slight variance, the average loss of the models is much lower than that corresponding to the convergence scenario. Thus, analyzing the performance of a GAN from its loss curves is a cumbersome task. This motivates the need for better measures to quantitatively evaluate and infer the learning of a GAN.

\section{Toy Game}

We present our experiment on a toy dataset where the Nash and non-Nash critical points can be easily visualized. This enables a clear understanding of the behavior of the two DG estimates. Consider a zero-sum game set in a 2-Dimensional space, between two agents - $x$ and $y$, guided by the following payoff function $f(x,y)$ well explored in optimization literature \cite{DBLP:journals/corr/abs-1901-00838}
\begin{equation}
     \underset{y}{\min} \ \underset{x}{\max} \ f(x,y) = e^{-0.01(x^2+y^2)}((0.3 x^2 +y)^2+(x+0.5y^2)^2)
    \label{eqn:toy_game}
\end{equation}
Figure \ref{fig:toy} depicts the contour plot of $f(.)$ together with its Nash (denoted as $A$ colored in green) and non-Nash (denoted as $B$ colored in red) critical points. In-order to estimate the duality gap at each of these points, we compute $M_1$ and $M_2$ by optimizing $f(.)$ w.r.t $x$ and $y$ respectively to obtain $x^w$ and $y^w$, using both vanilla and perturbed DG estimation processes. The optimization is performed using an Adam optimizer, with a learning rate of $5e-4$ over 500 iterations. The radius of the perturbation ball is set to 0.01. The true duality gap would be zero at $A$ and positive at $B$. We observe (Figure\ref{fig:toy}) that in the vicinity of $A$, $x^w$ and $y^w$ estimated using both vanilla and perturbed DG methods converge to $A$, and the estimated duality gap is very close to zero, as expected. However, in the vicinity of $B$, $x^w$ and $y^w$ estimated using the vanilla approach remains stuck at $B$, resulting in the duality gap estimate becoming zero. On the other hand, $x^w$ estimated using our approach can escape the non-Nash critical point and reach a local optimum, resulting in a positive duality gap estimate value as expected at a non-Nash critical point. An interesting observation here is that $y^w$ estimated using our approach also converges to $B$. This is expected because evaluating the hessian at $B$ (see theoretical verification), we see that $B$ is locally unstable only for $x$ and is a local optimum for $y$. Thus, the duality gap estimated by our approach is consistent with the expected values, in contrast to the vanilla approach.

\subsection{Theoretical Verification}

\begin{figure}
    \centering
     \includegraphics[width = 0.7\linewidth]{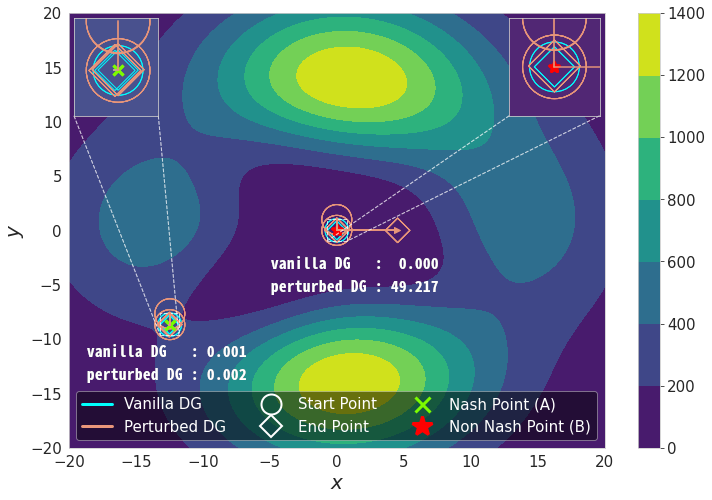}
       \caption{Toy Game - Duality Gap Estimation. The lines (see the zoomed insets) represent the optimization trajectories in finding $x^w$ and $y^w$. Circles denote the initial points of optimization and diamonds denote the optima reached, color coded according to the approach - orange represents our approach and cyan represent the vanilla approach.}
    \label{fig:toy}
\end{figure}

 We verify the dynamics of the mini-max game defined by Equation \ref{eqn:toy_game} in the vicinity of the critical points depicted in figure \ref{fig:toy}. \\

Let $A = (-12.43373,-8.78737)$ and $B = (0.0,0.0)$ .\\
We have,
\begin{center}
$f(x,y) = e^{-0.01(x^2+y^2)}((0.3 x^2 +y)^2+(x+0.5y^2)^2)$ \\
    
  
\end{center}


Evaluating the gradient and hessian at $B$,
\begin{center}

    $\nabla_{x}f(B) = 0.0 \And
    \nabla_{y}f(B) = 0.0$
    \\
    $\nabla^2_{xx}f(B) = 2.0 > 0 \And
    \nabla^2_{yy}f(B) = 2.0 > 0$
    \\

\end{center}
Thus $B$ is a local minima  w.r.t $y$, but is not a local maxima w.r.t $x$. Clearly, $B$ is a Non-Nash critical point. \\
Estimating the duality gap at $B$, we have :
\begin{center}
    Vanilla DG Estimate = 0.00 \\
    Perturbed DG Estimate = 49.217
\end{center}

Evaluating the gradients and hessian at $A$,
\begin{center}
    $\nabla_{x}f(A) =  0.06  \And
    \nabla_{y}f(A) = -0.06 $
    \\
    
    $\nabla^2_{xx}f(A) =  -1.1 <0 \And
    \nabla^2_{yy}f(A) =  9.8 >0$
    \\
    
\end{center}

Thus, $A$ is both a local minima  w.r.t $y$, and a local maxima w.r.t $x$. $A$ is thus a Nash critical point.
Estimating the duality gap at $A$, we have :
\begin{center}
    Vanilla DG Estimate = 0.001 \\
    Perturbed DG Estimate = 0.002
\end{center}

We thus verify that our approach is able to differentiate between convergence to Nash and Non Nash critical points more effectively than the vanilla duality gap estimation approach.

\section{Synthetic 2D datasets}
\begin{figure*}
\begin{center}
 \includegraphics[width = 0.85 \linewidth]{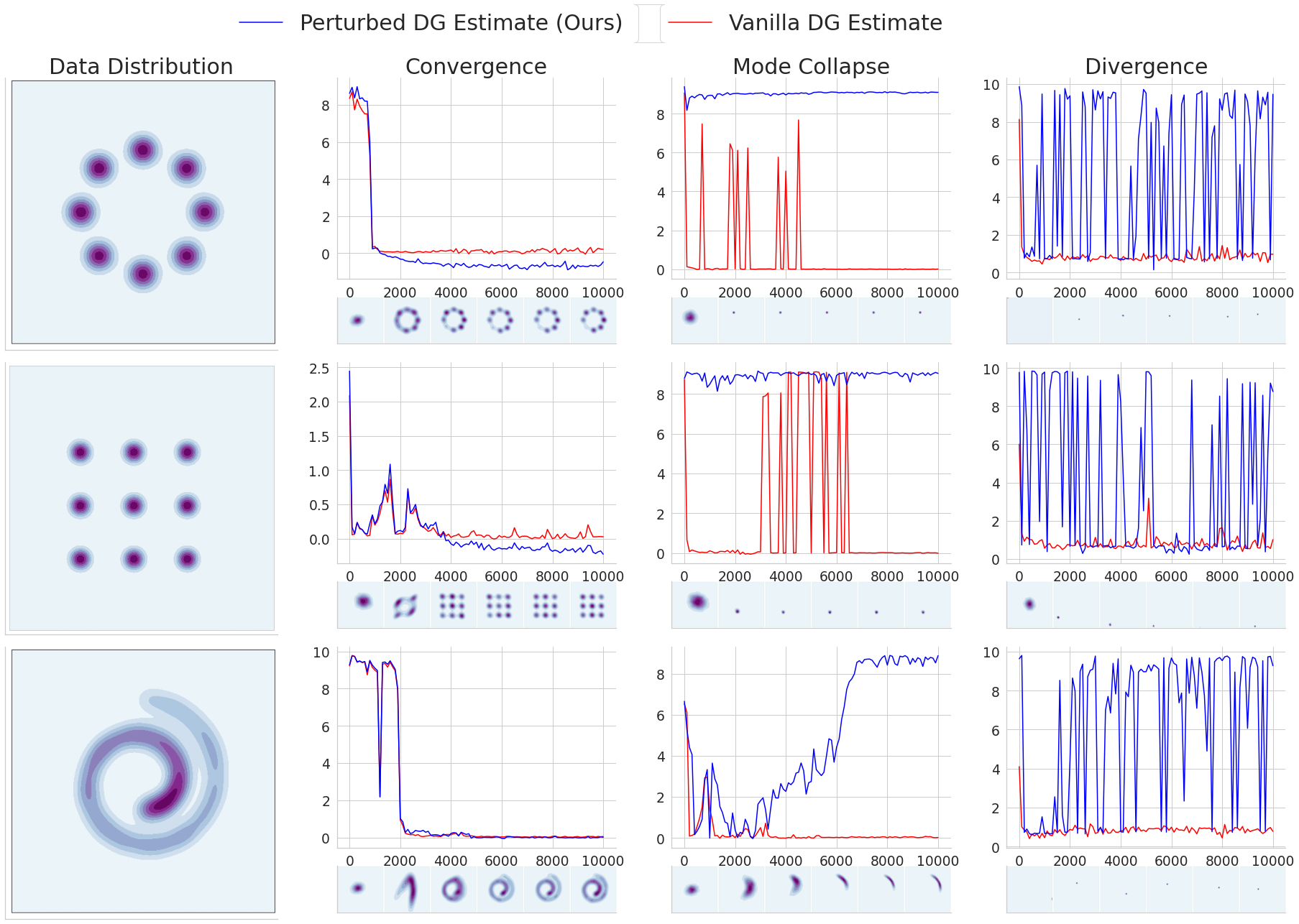}
\end{center}
  \caption{Perturbed and Vanilla DG estimate throughout the training progress of classic GAN on 2D data-sets for convergence, mode collapse, and divergence.}
\label{fig:2D-gan-grid}
\end{figure*}
   
\begin{figure*}
    \begin{center}
     \includegraphics[width = 0.85 \linewidth]{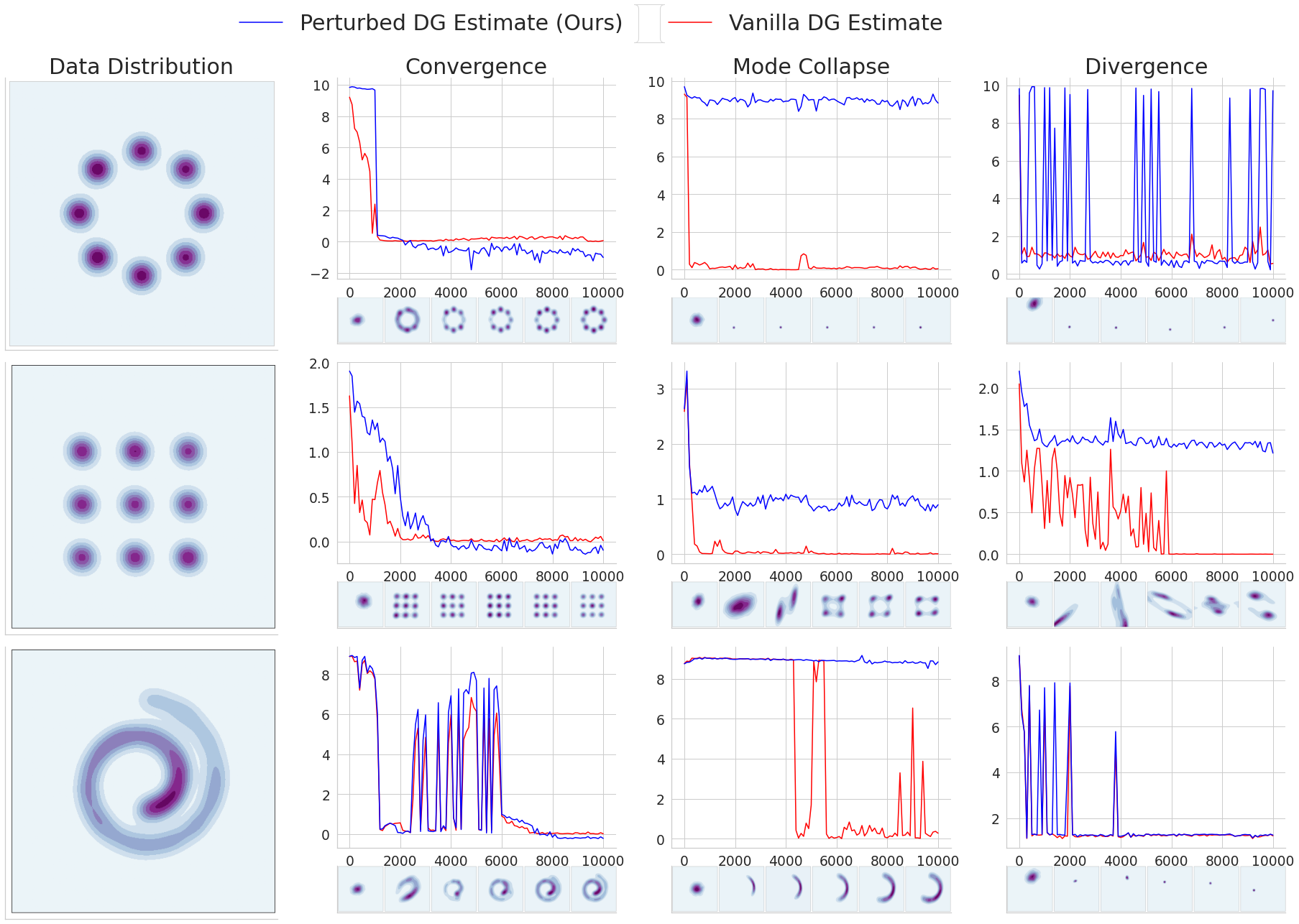}
    \end{center}
      \caption{Perturbed and Vanilla DG estimate throughout the training progress of NSGAN on 2D data-sets for convergence, mode collapse, and divergence.}
    \label{fig:2D-nsgan-grid}
\end{figure*}

The sensitivity of vanilla and perturbed DG to convergence and non-convergence scenarios during the training procedure of a classic GAN and a non-saturating (NS)GAN across the 2D synthetic datasets- RING, GRID and SPIRAL are presented in figures \ref{fig:2D-gan-grid} and \ref{fig:2D-nsgan-grid} respectively. We observe that almost in all cases of mode collapse and divergence, perturbed DG shows better sensitivity than vanilla DG by saturating (or oscillating) to a non-zero positive values.

 \subsection{Implementation Details}
 While training the classical GAN, we fix the learning rate as 5e-4 for both the generator and the discriminator across all scenarios. To simulate convergence, stable mode collapse and divergence, we use a discriminator-generator update ratio of 1:1, 15:1, and 1:15 respectively. While training NSGAN, we use learning rates as 1e-3, 1e-4, 1e-3, and update ratios as 3:2, 5:7, 1:10 respectively for convergence, mode collapse, and divergence scenarios. The learning rate is kept the same for both the generator and the discriminator. We use a latent noise dimension of $100$ for the generator. The architectural details of the models are summarized in Table-\ref{2D-model}. For the computation of the duality gap, we train the individual models using Adam optimizer for 300 iterations to find the worst-case generator/discriminator. The local perturbations added to the individual weight layers are drawn from a uniform distribution having standard deviation equal to twice the standard deviation of the weights of the corresponding layers.
 
\begin{table}
\begin{center}
\begin{tabular}{c c c c}
\hline
 & Generator & Discriminator   \\
\hline
& dense(128), relu & dense(128), relu \\
Architecture & dense(128), relu & dense(128), relu \\
& dense(2) & dense(1), sigmoid \\
\hline
optimizer & Adam & Adam \\
\hline

\end{tabular}
\end{center}
\caption{Model Details - 2D Experiments}

\label{2D-model}
\end{table}
 

\section{Image datasets}
  
    \begin{figure*}
    \begin{center}
     \includegraphics[width = 0.95\linewidth]{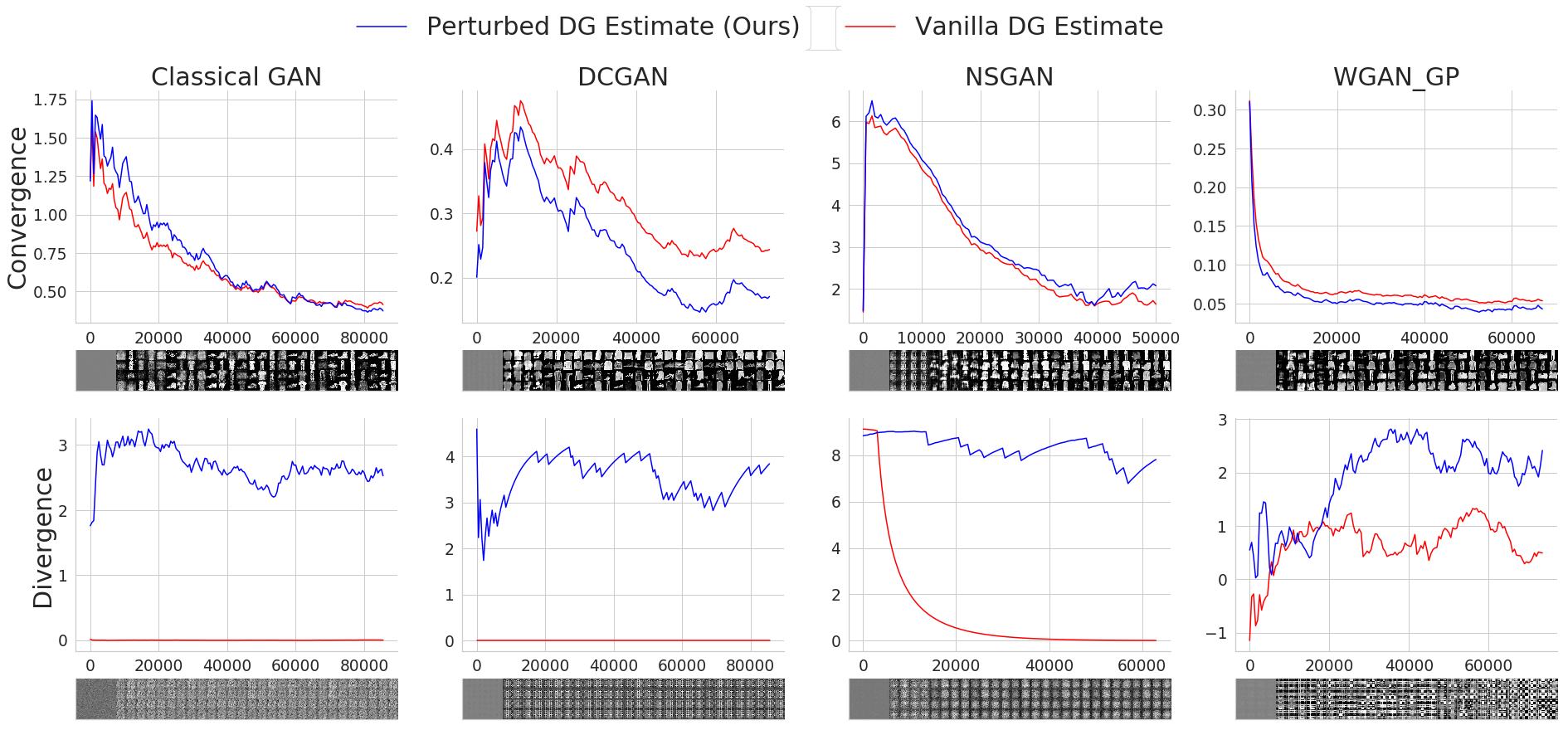}
    \end{center}
         \caption{Comparison of perturbed and vanilla DG estimate throughout the training progress of classic GAN (Col-1), DCGAN (Col-2), NSGAN (Col-3) and WGAN-GP (Col-4) on Fashion MNIST dataset for convergence (Row-1) and divergence (Row-2) settings. For each sub-graph, X-axis corresponds to the number of iterations, and Y-axis corresponds to DG values.}
    \label{fig:fashion}
    \end{figure*}
  
 \subsection{MNIST and Fashion MNIST}  
  \subsubsection{Convergence and Divergence on Fashion MNIST}   
  

  To show the generality of perturbed DG across datasets, we compare perturbed and vanilla DG estimates during the training progress of various GAN's on the Fashion MNIST dataset. Figure \ref{fig:fashion} shows the results obtained.
  We observe that perturbed DG is more sensitive to divergence setting than vanilla DG irrespective of the GAN type. Therefore our presumption that perturbed DG is robust across datasets and GAN architecture is valid.
\subsubsection{Implementation Details}
The architecture and hyper-parameter details of different GANs for convergence and divergence settings on MNIST and Fashion MNIST data-sets are discussed here. For classical GAN, we use fully connected layers followed by leaky relu activation with alpha = 0.3. To simulate convergence and divergence, we use 3 hidden layers with 512 nodes each, for both generator and discriminator, followed by a projection layer. For DCGAN, NSGAN, and WGAN-GP, we use convolutional layers followed by batch normalization and relu activation function. We project the latent noise to $7 \times 7 \times 256$ and use three transpose convolution layers containing 128, 64, 1 filters with strides 1, 2, 2, respectively for the generator.  For the discriminator, we use two convolution layers containing 64, 64 filters with strides 2, 2 respectively, followed by the projection layer. The convolutional  filter size is set to $5 \times 5$. The latent noise is of dimension 100. We fix the update ratio as 1:1 and use Adam optimizer across all settings for all GANs. The learning rates and the radius of the perturbation balls' (for computing DG) for convergence and divergence settings for different GANs on MNIST and Fashion MNIST datasets are given in Table \ref{gans-mnist-fashion-hyperparameters}(a) and \ref{gans-mnist-fashion-hyperparameters}(b) respectively.


\begin{table}[ht]
\centering
\subfloat[Subtable 1 list of tables text][]{
\begin{tabular}{c c c c c c}
        \hline
        & GAN\_Type & G\_lr & D\_lr & $\sigma$\\
        \hline
        & Classic GAN & 1e-4 & 1e-4 & 0.01 \\
        Convergence & DCGAN  & 2e-4 & 1e-4 & 0.01  \\
        & NSGAN & 2e-5 & 1e-5 & 0.01  \\
        & WGAN-GP & 5e-5 & 5e-5 & 0.2  \\
        \hline
        & Classic GAN & 2e-3 & 4e-3 & 0.1 \\
        Divergence & DCGAN & 1e-4 & 6e-4 & 0.1  \\
        & NSGAN & 7e-5 & 1e-5 & 0.05  \\
        & WGAN-GP & 1e-4 & 1e-2 & 0.5  \\
        \hline
        \end{tabular}}
\qquad
\qquad
\subfloat[Subtable 2 list of tables text][]{
\begin{tabular}{c c c c c c}
        \hline
        & GAN\_Type & G\_lr & D\_lr & $\sigma$\\
        \hline
        & Classic GAN & 1e-4 & 1e-4 & 0.01 \\
        Convergence & DCGAN  & 2e-4 & 1e-4 & 0.01  \\
        & NSGAN & 2e-5 & 1e-5 & 0.01  \\
        & WGAN-GP & 5e-5 & 5e-5 & 0.2  \\
        \hline
        & Classic GAN & 2e-3 & 4e-3 & 0.1 \\
        Divergence & DCGAN & 1e-4 & 6e-4 & 0.1  \\
        & NSGAN & 7e-5 & 1e-5 & 0.05  \\
        & WGAN-GP & 1e-4 & 1e-2 & 0.5  \\
        \hline
        \end{tabular}}
\caption{Hyper-parameter Details -(a) MNIST Experiments (b) Fashion MNIST Experiments}
\label{gans-mnist-fashion-hyperparameters}
\end{table}

 \subsubsection{Mode Collapse on MNIST and Fashion MNIST}
    
   \begin{figure}
    \begin{center}
     \includegraphics[width = 0.60 \linewidth]{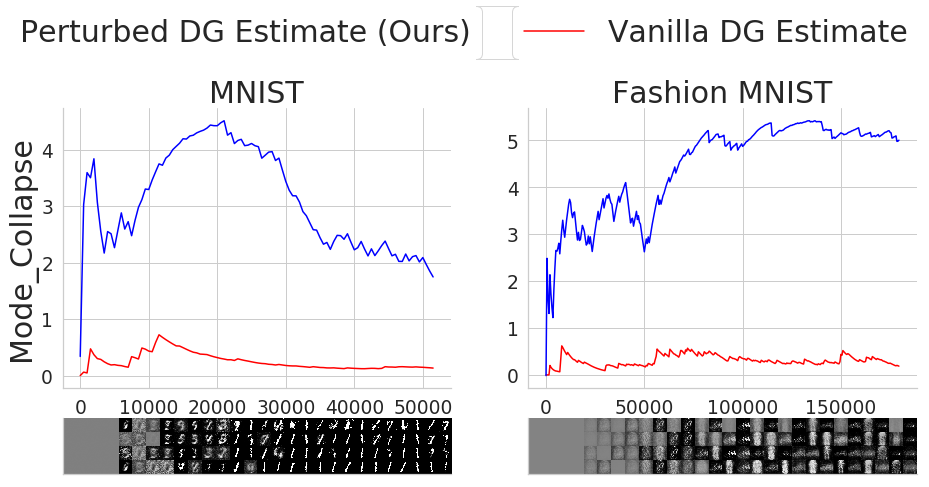}
    \end{center}
     \caption{Comparison of perturbed and vanilla DG estimate obtained during the training progress of classic GAN on MNIST (Col-1) and Fashion MNIST (Col-2) datasets for mode collapse setting.}
       
    \label{fig:mc_mnistFashion}
    \end{figure}
     

  In this experiment, our goal is to show that the sensitivity of the perturbed duality gap towards mode collapse setting is robust across datasets. It is challenging to simulate mode collapse in complex GAN architectures, so we choose classical GAN on MNIST and Fashion MNIST datasets for this experiment. We simulate mode collapse by tuning the hyperparameters (Table \ref{modecollapse-mnistfashion-hyperparameters}) and architecture of classic GAN. The architecture comprises of fully connected layers followed by leaky relu activation with alpha = 0.3. We use 5 hidden layers with 256 nodes each, for generator and 4 hidden layers with 256 nodes each, for the discriminator.

  The results of this experiment are illustrated in figure \ref{fig:mc_mnistFashion}. We observe that the perturbed DG is more sensitive to the mode collapse setting than vanilla DG as it saturates to a positive value, unlike vanilla DG, which is close to zero. 
 
  

  
   
\begin{table}
\begin{center}
\begin{tabular}{c c c c c c}
\hline
 & G\_lr & D\_lr & $\sigma$\\
\hline
MNIST &  1e-4 & 5e-4 & 0.1  \\

\hline
Fashion MNIST &  5e-5 & 5e-4 & 0.1  \\
\hline

\end{tabular}
\end{center}
\caption{Hyper-parameter details of classic GAN for mode collapse setting }

\label{modecollapse-mnistfashion-hyperparameters}
\end{table}
\begin{table}
\begin{center}
\begin{tabular}{c c c c c c c }
\hline
& Dataset & G\_lr & D\_lr  & $\sigma$ \\ 

\hline
& CIFAR10 & 1e-4 & 1e-4 & 1e-2 \\
Convergence & CelebA & 2e-4 & 1e-4 & 1e-2  \\
& CIFAR10 & 1e-4 & 1e-2 &  5e-1 \\
Divergence & CelebA & 5e-5 & 5e-1  & 5e-1  \\
\hline

\end{tabular}
\end{center}
\caption{Hyper-parameter Details - WGAN-GP on CIFAR10 and CelebA Experiments. }

\label{wgan-cifar-celeb-hyperparameters}
\end{table}


\subsection{ CIFAR-10 and CelebA }
\subsubsection{ Implementation Details }
We use WGAN-GP for conducting experiments on CIFAR-10 and CelebA datasets owing to its training stability. We project the latent noise to $4 \times 4 \times 256$ and use four transpose convolution layers containing 64, 64, 64, 3 filters with strides 2, 2, 1, 2 respectively for the generator.  For the discriminator, we use three convolution layers containing 64, 64, 64 filters with strides 2, 2, 2  respectively, followed by the projection layer. Also, a standard discriminator to generator update ratio of 5:1 is used. Other implementation details of WGAN-GP is the same as discussed under MNIST and Fashion MNIST datasets.  The hyper-parameter details are given in Table \ref{wgan-cifar-celeb-hyperparameters}.

\section{Dynamic Scheduling Using Perturbed DG}
Following the approach of \cite{xu2018autoloss}, we use reinforcement learning framework to train a controller to guide the GAN optimization. The state space comprises of a 4-tuple - (a) the log-ratio of the magnitude of the gradients of the generator and discriminator, an exponential moving average of - (b) generator's loss, (c) discriminator's loss, and (d) perturbed DG. The controller learns to output the posterior probability of choosing the agents (Generator and Discriminator) given the current state of optimization. The agent to be optimized at each iteration is sampled using the probability distribution defined by the contoller's output. Since such a sampling is non-differentiable, the parameters of the controller are learned using the REINFORCE \cite{10.1007/BF00992696} algorithm, where the reward for the controller's action is defined as the downstream GAN performance in terms of perturbed DG. More specifically, we define the reward as $\frac{\alpha}{DG + \epsilon}$ where $DG$ is the exponential moving average of perturbed DG at the end of a training episode and $\epsilon$ is added for numerical stability. An episode consists of $K$ optimization steps for the GAN, guided by the controller. On termination of an episode, the expected reward corresponding to each action in the dynamic schedule generated by the controller is computed and the controller is updated to maximize this reward. An episode may also terminate if the output of the controller collapses to the same action, during which a negative reward is provided.

\begin{figure}
\centering
\begin{tabular}{cc}
\subfloat[GRID Dataset -Fixed Ratio 5:1]{\includegraphics[width = 0.45\linewidth]{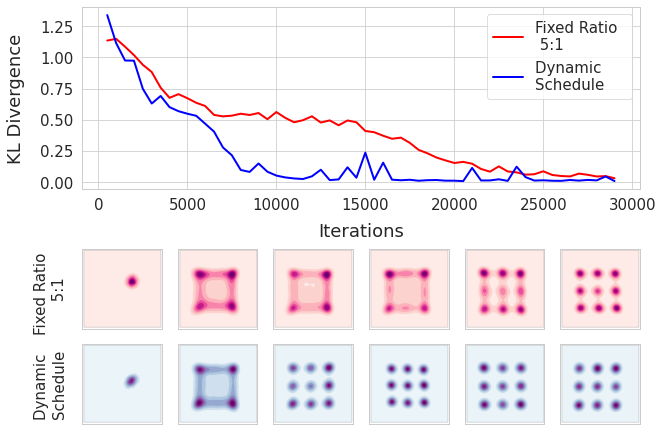}} &
\subfloat[SPIRAL Dataset - Fixed Ratio 5:1]{\includegraphics[width = 0.45\linewidth]{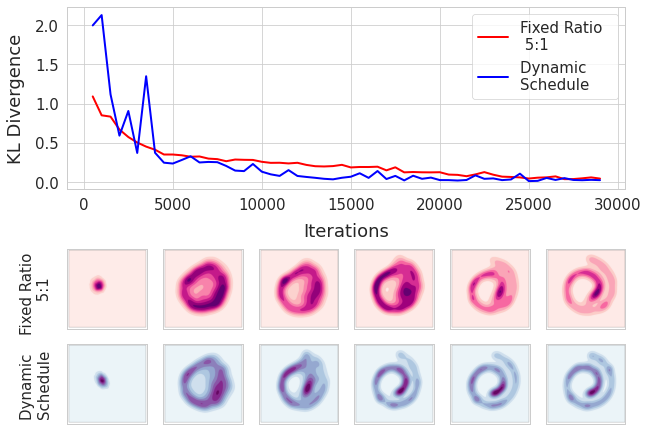}} \\
\subfloat[GRID Dataset -Fixed Ratio 1:5]{\includegraphics[width = 0.45\linewidth]{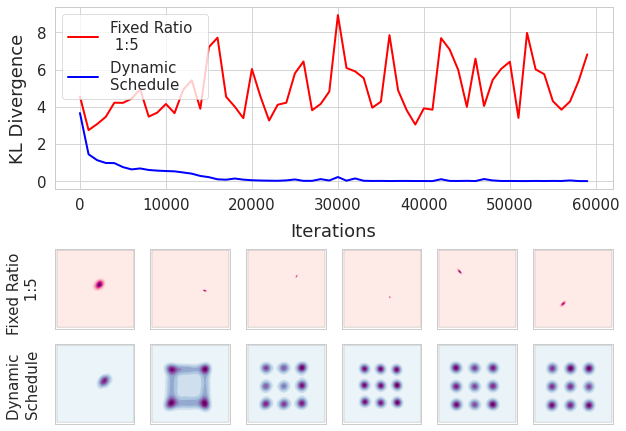}} &
\subfloat[SPIRAL Dataset -Fixed Ratio 1:5]{\includegraphics[width = 0.45\linewidth]{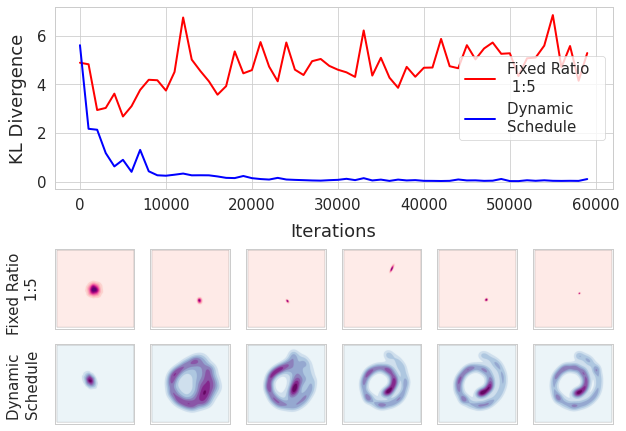}} \\
\end{tabular}
\caption{Dynamic scheduling using perturbed DG leads to faster and better convergence}
\label{fig:dynamic_schedule_sp}
\end{figure}

We train the controller to guide the optimization of a WGAN-GP on the RING dataset. We test this controller on the GRID and SPIRAL datasets. Figure \ref{fig:dynamic_schedule_sp} compares the performance of the GAN using the dynamic schedule given by the controller as opposed to a fixed schedule. We consider two fixed (D:G) update ratios - 5:1 and 1:5. We observe that though the standard 5:1 fixed ratio eventually leads to convergence for both GRID and SPIRAL, using the dynamic schedule leads to faster convergence. This is evident from the KL divergence plots and the visualization of the generated distribution across iterations. On the other hand, using a fixed update ratio of 1:5 leads to divergence. Thus, the controller trained using perturbed DG helps supersede the effort that manual tuning of the update ratio demands.

\begin{figure*}
    \includegraphics[width = \linewidth]{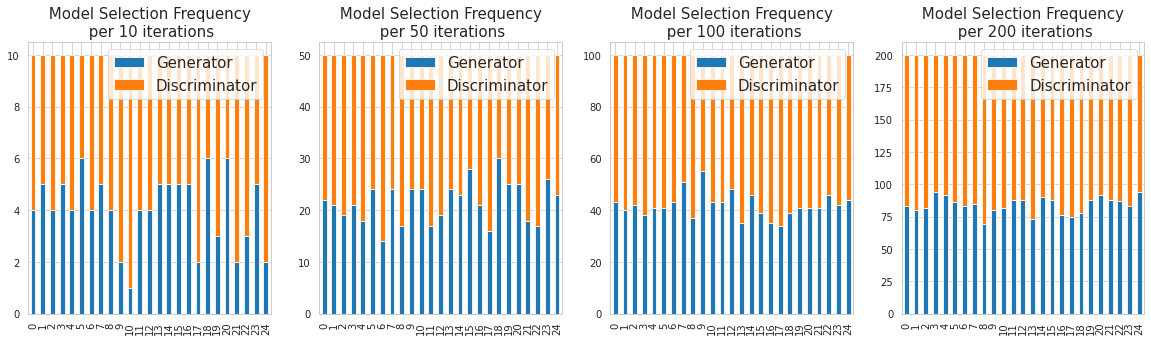}
      \caption{Dynamic Schedule : Selection Frequency of Generator and Discriminator within intervals of varying resolution. Across the x-axis are successive intervals of the specified resolution. The Y axis depicts the proportion of times each model is selected within the specified interval.}
    \label{fig:selection_frequency}
\end{figure*}

Figure \ref{fig:selection_frequency} represents the selection frequency of the generator and discriminator using the dynamic schedule within intervals of varying resolution across the training iterations of a WGAN-GP on the GRID dataset. We observe that the variance in the selection frequencies is lower within intervals of higher resolution, with the discriminator chosen slightly more than the generator on an average. This is consistent with the standard suggestion for WGAN-GP that the discriminator be given larger updates than the generator and implies that there is an overall balance that is maintained. However, the dynamic nature of the updates that is visible within intervals of lower resolution emphasize that the delicate balance between the agents is dependent on the state of the optimization. An agent's optimal response to its adversary's current strategy might require multiple optimization steps. Thus, the dynamic schedule parameterized by the controller and learned using perturbed DG, is a better candidate over a fixed schedule to enforce faster convergence for GANs.

\subsection{Implementation Details}
The controller network consists of two dense layers with 128 and 64 nodes respectively, each using a tanh activation, followed by the output projection layer having two nodes with softmax activation. We use an Adam optimizer with an initial learning rate of 1e-3 to update the paramaters of the controller. We specify an episode to consist of $K$ = 30000 iterations, every 500 of which we compute perturbed DG and maintain the exponential moving average to be used in the reward computation. The value of the reward constant $\alpha$ is set to 5 and $\epsilon$ is set to 1e-5 for numerical stability. As for the task model (WGAN-GP) on 2D synthetic datasets, we use 3 dense layers with 128 nodes each having relu activation, followed by the projection layer, for both the generator and discriminator. To test for MNIST, we follow the same architecture of WGAN-GP discussed under the Image Datasets section. The dimension of the latent space is set to 100 for all tasks. The parameters of the task model are updated using an Adam optimizer with an initial learning rate of 5e-4. To compute perturbed DG, the auxilary models are optimized for 300 iterations, with the radius of the perturbation ball set to 0.25.

\section{Ablation Studies}
\subsection{Trend of Perturbed DG across Training Iterations of GAN}

\begin{figure}
\begin{center}
 \includegraphics[width = 1.0\linewidth]{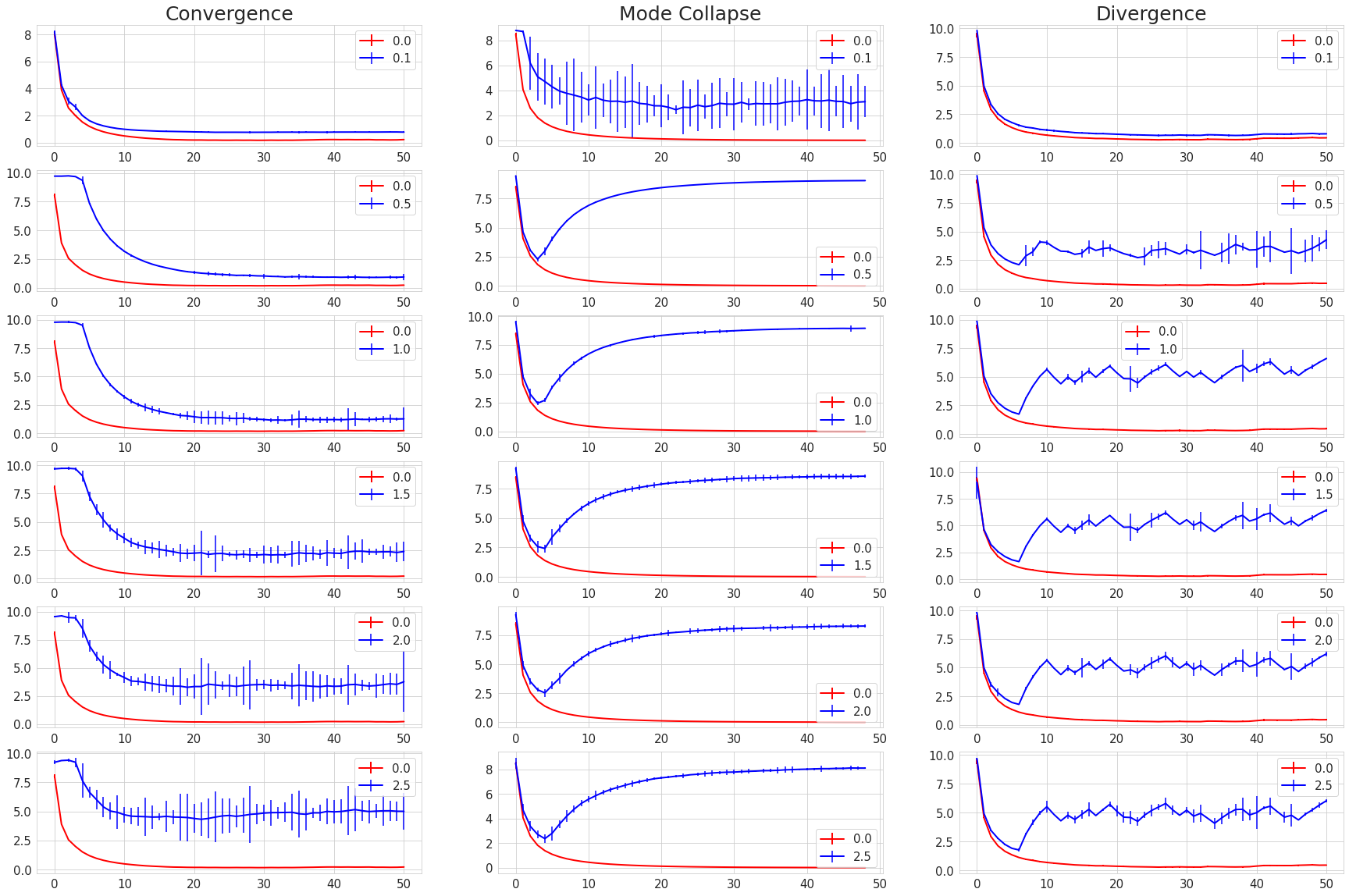}
\end{center}
   \caption{Trend of perturbed (blue) and vanilla (Red) DG across GAN training iterations. Rows correspond to different $\sigma$ values and columns correspond to settings. Legends in each subplot represent the radius of the perturbation ball corresponding to the curves. Curves corresponding to perturbation ball of radius zero is equivalent to the vanilla DG estimate. Within each subplot, the y-axis corresponds to the estimated duality gap and x-axis corresponds to the training step. The x-axis scale is in 200 iterations.}
\label{fig:ablation_trend}
\end{figure}

We monitor the trend of perturbed DG (blue) across training iterations w.r.t to radii of perturbation ball $\sigma$ for convergence, mode collapse, and divergence settings. For each experiment, we train a classic GAN for 10000 iterations, and the average plots across 10 trials are shown in figure \ref{fig:ablation_trend}. As we move from top to bottom, $\sigma$ increases. For the convergence setting (first column), perturbed DG values increase gradually with $ \sigma $. However, in mode collapse and divergence setting (second and third column), perturbed DG increases abruptly and then saturates. This result is consistent with figure 5(a) of the main paper,
where we show that perturbed DG increases gradually with $\sigma$ for convergence setting but saturates immediately after a sharp increase in case of mode collapse and divergence settings. 
However, there we had analyzed the effect of $\sigma$ at the end of the training process; here, analysis is done during training of GAN.

\subsection{Trend of $\boldsymbol{M_1} \mbox{and} \boldsymbol{M_2}$ across Training Iterations of Auxiliary GAN} 
    
    \begin{figure*}
        \centering
        \begin{tabular}{cc}
        \subfloat[$M_1$ Vs Iterations ]{\includegraphics[width = 0.5\linewidth]{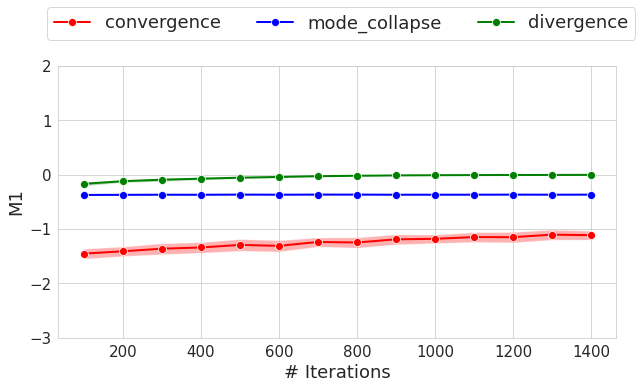}} &
        \subfloat[$M_2$ Vs Iterations]{\includegraphics[width = 0.5\linewidth]{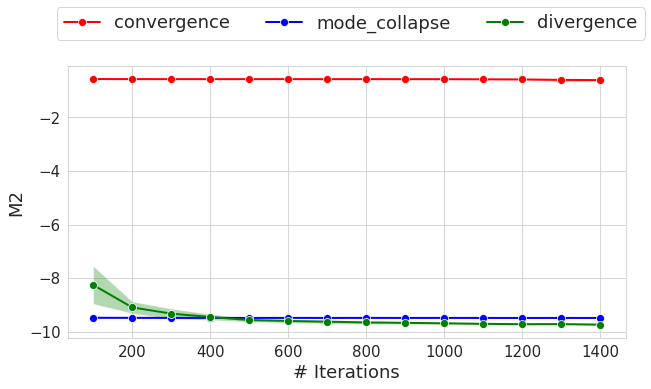}} \\
        \end{tabular}
        \caption{Trend of $M_1$ and $M_2$ w.r.t training iterations of an auxiliary GAN for convergence, mode collapse and divergence settings.}
        \label{fig:dg-early-stopping}
    \end{figure*}
    We study the effect of the number of iterations involved in estimating $M_1$ and $M_2$ using our approach. We train a classic GAN for 10000 iterations across the three different settings and present the variation in the final $M_1$ and $M_2$ estimates w.r.t increase in the number of iterations used in the optimization of the auxiliary models. In figure \ref{fig:dg-early-stopping}(a), we observe that there is no significant change to $M_1$ post 200 iterations for mode collapse and divergence settings. However, $M_1$ gradually increases and saturates for the convergence setting. A possible reason for this is that the discriminator is initially confused during convergence and optimizing it, keeping the generator fixed, enables it to learn a more discriminative decision boundary. However, the slope being marginal reflects that it is within the influence of the Nash point despite the perturbation. 
    
    In figure \ref{fig:dg-early-stopping}(b) we observe that there is no significant change to $M_2$ post 300 iterations for all the three settings. The $M_1$ and $M_2$ values being close during convergence further verifies that the perturbation does not displace the agents from the influence of the Nash point. 

\bibliographystyle{unsrt}  
\bibliography{bibfile}